\documentclass[runningheads]{llncs}
\usepackage[T1]{fontenc}
\usepackage{amsmath} 
\usepackage{booktabs}
\usepackage{array}
\usepackage{multirow}
\usepackage{hyperref}
\usepackage{graphicx,verbatim}
\begin{document}
\title{Understanding Sources of Demographic Predictability in Brain MRI via Disentangling Anatomy and Contrast}
\titlerunning{Demographic predictability in Brain MRI}
%
\author{Mehmet Yigit Avci$^{*,\ddagger}$\inst{1} \and
Akshit Achara$^{*}$\inst{1} \and
Andrew P. King$^{\dagger}$ \inst{1}\and
Jorge Cardoso$^{\dagger}$\inst{1,2},
for the Alzheimer’s Disease Neuroimaging Initiative}


\authorrunning{Avci et al.}
%
\institute{King's College London, UK
\email{\{yigit.avci,akshit.achara,m.jorge.cardoso,andrew.king\}@kcl.ac.uk}}
  
\maketitle              
\let\thefootnote\relax\footnotetext{* These authors contributed equally to this work.}
\let\thefootnote\relax\footnotetext{$\dagger$ These authors jointly supervised this work.}
\let\thefootnote\relax\footnotetext{$\ddagger$ Corresponding author: \texttt{yigit.avci@kcl.ac.uk}.}
\begin{abstract}
Demographic attributes can be predicted from medical images, raising concerns about bias in clinical AI systems. In X-ray imaging, acquisition characteristics have been shown to contribute substantially to this predictability. Whether the same holds in brain MRI remains unclear, as anatomical variation and acquisition-dependent contrast are deeply entangled in the image formation process, obscuring the origins of demographic signal. To address this, we propose a controlled framework based on disentangled representation learning, decomposing brain MRI into anatomy-focused representations that suppress acquisition influence and contrast embeddings that capture acquisition-dependent characteristics. Training predictive models for age, sex, and race on full images, anatomical representations, and contrast embeddings allows us to quantify the relative contributions of structure and acquisition to the demographic signal. Across three datasets and multiple MRI sequences, demographic predictability is found to be driven primarily by anatomical variation, with anatomy-focused representations largely preserving the performance of models trained on raw images. Contrast embeddings retain a weaker signal that is dataset-specific and does not generalize across sites. These findings suggest that effective mitigation must explicitly account for the primarily anatomical and secondarily acquisition-dependent origins of demographic signal, ensuring that any bias reduction generalizes robustly across domains.


\end{abstract}
\section{Introduction}

Medical imaging is essential for clinical practice and biomedical research, supporting diagnosis, treatment planning, and disease monitoring. While machine learning has enabled powerful image-based models for tasks such as segmentation, classification, and prognosis, growing evidence suggests that these models can exploit unintended correlations in imaging data, leading to performance disparities across demographic groups and allowing attributes such as race to remain identifiable even when unrelated to the clinical task~\cite{Underdiagnosis_2021,larrazabal2020,biasinrad,esthercmrfairness,gichoya2022ai}. Attempts to mitigate these effects through generic fairness interventions have proven insufficient, often degrading performance or failing under real-world distribution shifts~\cite{schrouff2023diagnosingfailuresfairnesstransfer}. These observations suggest that the demographic signal is not a superficial modelling artifact but reflects structured variation embedded within the data itself.

Recent work has begun to investigate where this signal originates. For example, in cardiac MRI segmentation, demographic information has been shown to persist within learned representations despite controlled task settings~\cite{Lee_2025}. In chest X-ray classification, acquisition-related factors substantially contribute to the demographic signal and downstream performance disparities~\cite{lotter2024acquisition}. Collectively, these findings indicate that bias may also emerge from modality and acquisition-specific characteristics that models implicitly exploit.

This perspective is particularly relevant for brain MRI. Although demographic bias has been documented in MRI-based models~\cite{achara2025invisibleattributesvisiblebiases,biasmri2022,melba:2025:035:danaee}, its origins remain unclear. Brain MRI is intrinsically heterogeneous as image appearance is governed by interactions between sequence parameters (e.g., repetition time (TR), echo time (TE)), scanner hardware, and site-specific protocols. These acquisition-dependent factors systematically shape intensity distributions and tissue contrast, introducing variability sources that correlate with demographic attributes. As anatomical structure and acquisition-dependent contrast are entangled in raw images, conventional models cannot determine whether the demographic signal arises from biological variation, acquisition settings, or their interaction. 

To understand the underlying drivers of demographic bias in brain MRI, we employ disentangled representation learning frameworks, specifically MR-CLIP~\cite{avci2025metadataaligned3dmrirepresentations} and DIST-CLIP~\cite{avci2025distcliparbitrarymetadataimage}. These models decompose brain MRI into distinct anatomical images and contrast embeddings, enabling explicit separation of structural information from acquisition-dependent signal. We train predictive models for age, sex, and race using (i) full brain MRI images, (ii) anatomy-focused images that suppress acquisition effects, and (iii) contrast embeddings that capture acquisition characteristics while minimizing anatomical content. By comparing predictive performance across these representations, we quantify the relative contributions of anatomical structure and acquisition-induced contrast to the demographic signal. Our contributions are: (1)~a controlled framework for disentangling anatomical and acquisition-dependent sources of the demographic signal in brain MRI; (2)~evidence, across three datasets and multiple MRI sequences, that anatomical structure is the dominant carrier of demographic information; and (3)~showing that contrast embeddings retain only a weak, dataset-specific signal that does not generalize across sites, establishing that effective bias mitigation strategies must primarily address anatomical variation. Depending on site-specific demographics and protocols, acquisition characteristics may also require consideration.
\section{Methods}

\textbf{Experimental Design.} Let $\mathcal{D} = \{(x_i, y_i, a_i)\}_{i=1}^{N}$ denote a dataset of brain MRI scans $x_i \in \mathcal{X}$, demographic attributes $y_i \in \mathcal{Y}$ (age, sex, race), and acquisition metadata $a_i \in \mathcal{A}$ (e.g., TR, TE, scanner type, site). Our objective is to determine whether the demographic signal arises primarily from anatomical structure, acquisition-dependent contrast, or their interaction.

In raw MRI, anatomical structure and contrast are inherently entangled within the image space $\mathcal{X}$. We therefore seek a representation-level decomposition of each image $x$ into two components, 
\begin{equation}
x \;\longrightarrow\; (z_{\text{anat}}, z_{\text{contrast}}), 
\end{equation}
where $z_{\text{anat}} \in \mathcal{Z}_{\text{anat}}$ captures structural information while minimizing acquisition effects, and $z_{\text{contrast}} \in \mathcal{Z}_{\text{contrast}}$ captures acquisition-dependent signal while reducing anatomical dominance.  To operationalize this separation, we leverage pre-trained disentangled representation learning frameworks, MR-CLIP~\cite{avci2025metadataaligned3dmrirepresentations} and DIST-CLIP~\cite{avci2025distcliparbitrarymetadataimage}.
MR-CLIP is a metadata-guided contrastive learning framework that aligns MRI images with their associated DICOM acquisition parameters without requiring manual sequence labels. An image encoder $E_I$ and a metadata encoder $E_M$ are trained jointly so that the image embedding $z_{\text{contrast}}^x = E_I(x)$ and the metadata embedding $z_{\text{metadata}}^x = E_M(a)$ are similar for matched pairs and dissimilar for mismatches. By optimizing a contrastive objective over image–metadata pairs, MR-CLIP learns a 512-dimensional embedding that reflects acquisition differences~\cite{avci2026mrclip,avci2025metadataaligned3dmrirepresentations}. DIST-CLIP builds upon MR-CLIP by introducing an explicit anatomical pathway to separate structural content from acquisition-dependent contrast. It employs an anatomy mapper that projects an input image $x$ to an anatomy-focused representation $z_{\text{anat}}$. This representation is designed to preserve geometric and morphological structure while suppressing intensity and contrast-related cues. To enforce acquisition invariance, DIST-CLIP incorporates contrastive loss applied at the patch level across anatomy images. This loss encourages corresponding anatomical patches from images acquired with different protocols to remain similar in the $z_{\text{anat}}$ space, regardless of acquisition differences. This explicitly promotes structural consistency while discouraging the encoding of contrast-specific information. As a result, the model achieves a disentangled decomposition in which $z_{\text{anat}}$ captures acquisition-invariant information, while $z_{\text{contrast}}$ encodes acquisition-dependent characteristics ~\cite{avci2025distcliparbitrarymetadataimage}.

\begin{figure}[htpb!]
\centering
\resizebox{.9\textwidth}{!}{
\includegraphics[width=\linewidth]{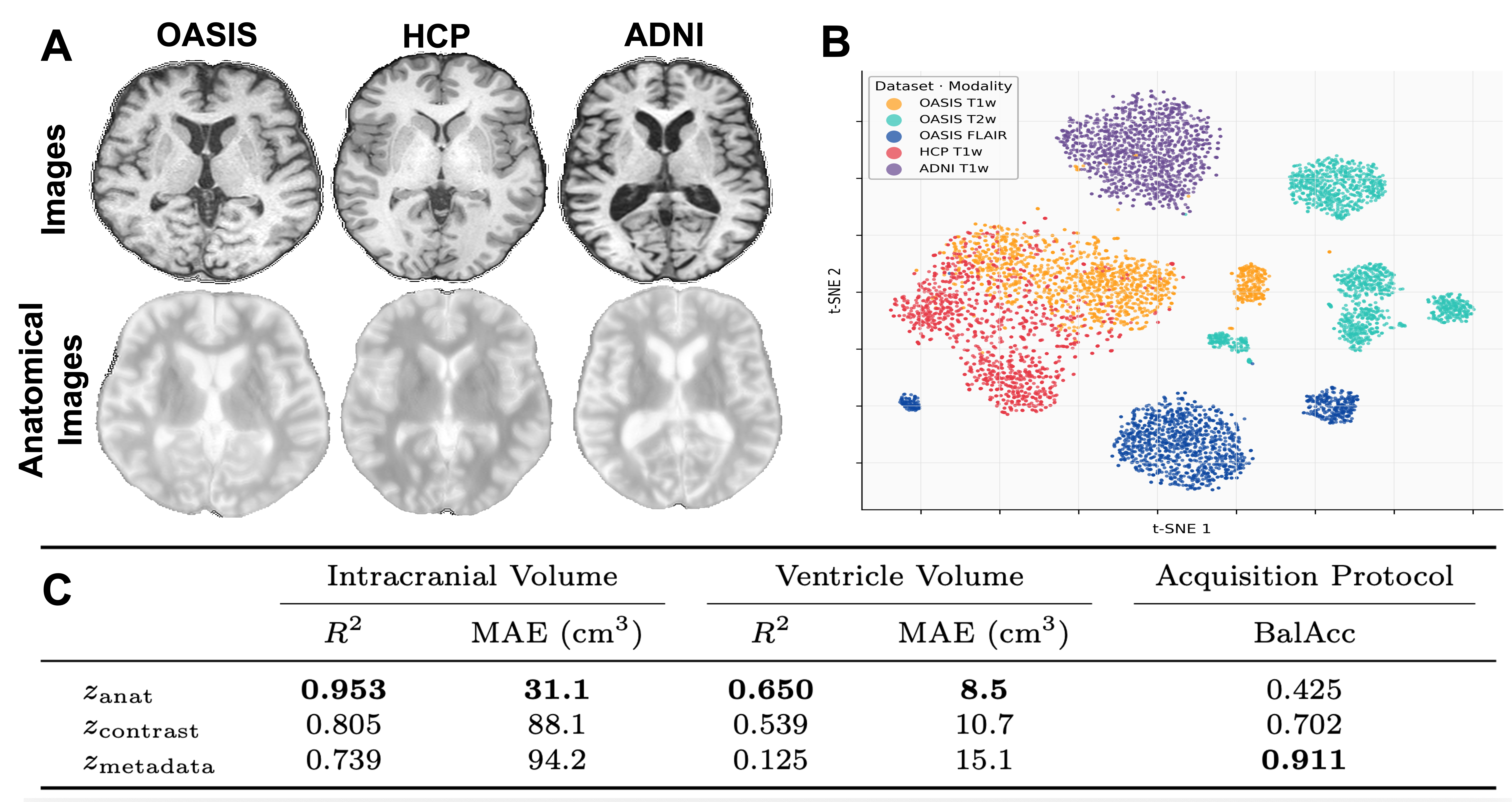}}
\caption{
Qualitative and embedding-level visualization of the disentanglement.
\textbf{A}: Raw images ($x$, top) and anatomy-focused representations ($z_{\text{anat}}$, bottom).
\textbf{B}: t-SNE of contrast embeddings ($z_{\text{contrast}}$) colored by dataset and sequence.
\textbf{C}: Linear probing ($N\!=\!500$). Anatomical probes predict intracranial volume (ICV) and ventricle volume; the acquisition probe predicts scanner/protocol cluster (15 classes; chance\,=\,6.7\%). $z_{\text{metadata}}$ denotes the metadata-only embedding from $E_M$.
}
\label{fig:disentanglement}
\end{figure}

Fig.~\ref{fig:disentanglement} provides qualitative and quantitative evidence of the representation separation. Panel~A shows raw T1-weighted images from OASIS, HCP, and ADNI alongside their anatomy-focused counterparts ($z_{\text{anat}}$): while raw images vary visibly in intensity and tissue contrast across sites, the anatomical representations retain structural detail and suppress these acquisition-driven differences. Panel~B visualizes contrast embeddings ($z_{\text{contrast}}$) via t-SNE, where clear clustering by dataset and MRI sequence confirms that acquisition-dependent characteristics are well captured. The framework does not assume perfect disentanglement, an inherently unattainable objective, but instead we explicitly quantify the degree of disentanglement through a probing analysis (Panel C). Using SynthSeg~\cite{synthseg}, we extract ground-truth intracranial volume (ICV) and ventricle volume for $N\!=\!500$ OASIS-3 \cite{lamontagne2019oasis3} subjects, then train linear regression and classification probes on dimensionally matched representations under Group K-Fold cross-validation. The results show that $z_{\text{anat}}$ strongly predicts anatomical features (e.g., ICV $R^2\!=\!0.953$, MAE\,=\,31.1\,cm$^3$) but poorly predicts acquisition protocol (BalAcc 42.5\%, where chance is 6.7\% over 15 scanner/protocol clusters). Conversely, $z_{\text{contrast}}$ is informative about acquisition identity (BalAcc 70.2\%) while encoding substantially less anatomical detail (ICV $R^2\!=\!0.805$, MAE\,=\,88.1\,cm$^3$). Crucially, the metadata-only embedding $z_{\text{metadata}}$, derived from DICOM parameters with no access to the image, already achieves (ICV $R^2\!=\!0.739$, MAE\,=\,94.2\,cm$^3$), indicating that most of the residual anatomical signal in $z_{\text{contrast}}$ does not reflect true image-level leakage, but rather population-level correlations between acquisition settings and demographic composition across sites.

To quantify the contribution of each representation to the demographic signal, we define three predictive mappings:
\begin{align*}
f_{\text{full}} &: \mathcal{X} \rightarrow \mathcal{Y}, \\
f_{\text{anat}} &: \mathcal{Z}_{\text{anat}} \rightarrow \mathcal{Y}, \\
f_{\text{contrast}} &: \mathcal{Z}_{\text{contrast}} \rightarrow \mathcal{Y}.
\end{align*}

Here, $f_{\text{full}}$ is trained directly on raw MRI $x$, while $f_{\text{anat}}$ and $f_{\text{contrast}}$ operate on disentangled representations $z_{\text{anat}}$ and $z_{\text{contrast}}$, respectively. 
For each demographic attribute $y \in \{\text{age}, \text{sex}, \text{race}\}$, we train models under identical optimization protocols and compare predictive performance across these mappings. Performance differences between $f_{\text{anat}}$ and $f_{\text{contrast}}$ provide an empirical estimate of the relative contribution of anatomical structure and acquisition-dependent contrast to the demographic signal. The performance of $f_{\text{full}}$ serves as a reference for the total demographic information accessible from the raw image $x$.

\textbf{Models and Evaluation.} For image-based inputs ($f_{\text{full}}$ and $f_{\text{anat}}$), we employ a 3D ResNet-50 architecture \cite{he2015deepresiduallearningimage,solovyev20223d} operating on volumetric inputs of size $128 \times 128 \times 128$. All models are initialized from ImageNet-pretrained weights (inflated to 3D). A fixed learning rate of $1\times10^{-4}$ is used across all experiments with dropout regularization of 0.2 to mitigate overfitting. The batch size is 2 due to memory constraints of 3D training. For embedding-based inputs ($f_{\text{contrast}}$), we use a multi-layer perceptron operating on $z_{\text{contrast}}$ with architecture
512 (input) $\rightarrow$ 512 $\rightarrow$ 256 $\rightarrow$ $N_{\text{classes}}$,
with ReLU activations and dropout (0.2) applied after each hidden layer.

Age prediction is formulated as a regression task optimized using mean squared error (MSE) (HCP age uses range midpoint). Sex and race prediction are formulated as classification tasks optimized with cross-entropy loss. All models are trained under matched data splits, learning rates, and regularization schemes to ensure comparability across representation types. We report mean absolute error (MAE) for age prediction and balanced accuracy (BalAcc) for classification tasks, ensuring equal weighting of classes to account for potential imbalance. Balanced accuracy is defined as the average recall across classes (i.e., mean per-class sensitivity), thereby preventing inflated performance estimates in the presence of class imbalance.

\textbf{Datasets and Preprocessing.} We evaluate our approach on three publicly available brain MRI datasets: Open Access Series of Imaging Studies (OASIS)~\cite{lamontagne2019oasis3}, Alzheimer’s Disease Neuroimaging Initiative (ADNI)~\cite{jack2008adni_mri}, and Human Connectome Project (HCP)~\cite{hcp}. These cohorts span different age ranges, scanner vendors, and acquisition protocols, providing diverse combinations of $(x_i, y_i, a_i)$. This heterogeneity is essential for disentangling anatomical structure from contrast variations induced by imaging parameters. The demographic distribution of each dataset, including age, sex, and race, is summarized in Table~\ref{tab:demographics}.
To ensure consistency across datasets and compatibility with the disentanglement framework, all volumes are rigidly registered to the MNI152 atlas at 1.0\,mm$^3$ isotropic resolution. Skull stripping is performed using SynthStrip~\cite{synthstrip}. 

\begin{table}[htb]
\centering
\caption{Subject-level demographic summary of the included datasets. Age is reported as mean $\pm$ standard deviation. Sex and race are reported as counts (percentage within dataset).}
\label{tab:demographics}
\setlength{\tabcolsep}{5pt}
\renewcommand{\arraystretch}{1.15}
\resizebox{\textwidth}{!}{
\begin{tabular}{l c  c c  c c}
\toprule
\multirow{2}{*}{\textbf{Dataset}} 
& \multicolumn{1}{c}{\textbf{Age}} 
& \multicolumn{2}{c}{\textbf{Sex}} 
& \multicolumn{2}{c}{\textbf{Race}} \\
\cmidrule(lr){2-2} 
\cmidrule(lr){3-4} 
\cmidrule(lr){5-6}
& Mean $\pm$ SD 
& Female 
& Male 
& White 
& Black
\\
\midrule
OASIS-3 & 72.4 $\pm$ 9.1 & 608 (55.6\%) & 485 (44.4\%) & 926 (84.7\%) & 167 (15.3\%) \\
ADNI & 74.7 $\pm$ 7.3 & 725 (44.6\%) & 901 (55.4\%) & 1542 (94.8\%) & 84 (5.2\%) \\
HCP & 29.1 $\pm$ 3.5 & 543 (54.4\%) & 455 (45.6\%) & 830 (83.2\%) & 168 (16.8\%) \\
\bottomrule
\end{tabular}}
\end{table}

\section{Experiments and Results}

To examine how anatomical and acquisition-dependent information contribute to demographic predictability, we compare the signal available from each representation within individual datasets and under joint training, test whether this predictability transfers across datasets as cohort and acquisition characteristics change, and assess whether the same anatomy–contrast pattern holds across MRI sequences.

The within-dataset and joint-training results for T1-weighted images are summarized in Table~\ref{tab:auc_results}. Models trained on raw images achieve the highest overall performance across tasks (e.g., sex BalAcc 0.92–0.96; age MAE 2.67–5.15 years). Anatomy-focused representations largely preserve this performance, for instance, sex prediction reaches 0.93 in OASIS, indicating that substantial demographic signal is embedded in structural variation. When training jointly across all datasets, this pattern remains consistent (Sex 0.96 Raw vs. 0.96 Anat vs. 0.78 Contrast), suggesting that the dominance of anatomical signal is not driven by single-dataset effects. Importantly, contrast embeddings, despite explicitly minimizing anatomical content, retain non-trivial predictive power. Sex prediction from $z_{\text{contrast}}$ achieves BalAcc 0.71 in OASIS and 0.81 in HCP (0.78 in joint training), while race prediction remains above chance in OASIS (0.71), and joint training (0.68). Even for age, contrast embeddings yield informative MAE values (e.g., 6.28 years in OASIS, and 5.21 jointly). Although consistently lower than those of the raw-image and anatomy-based models, these results demonstrate that contrast alone retains a demographic signal when evaluated within a single domain.

\begin{table*}[htb]
\centering
\caption{Within-dataset and joint-training performance across representations. 
For each dataset (OASIS, ADNI, HCP) and joint training across all datasets (ALL), 
we report balanced accuracy (BalAcc, $\uparrow$; equal class weighting) for \textbf{Sex} and \textbf{Race}, 
and mean absolute error (MAE, $\downarrow$) for \textbf{Age}. }
\label{tab:auc_results}
\setlength{\tabcolsep}{6pt}
\renewcommand{\arraystretch}{1.15}
\resizebox{\textwidth}{!}{
\begin{tabular}{l ccc ccc ccc}
\toprule
& \multicolumn{3}{c}{\textbf{Sex (BalAcc $\uparrow$)}} & \multicolumn{3}{c}{\textbf{Race (BalAcc $\uparrow$)}} & \multicolumn{3}{c}{\textbf{Age (MAE $\downarrow$)}} \\
\cmidrule(lr){2-4} \cmidrule(lr){5-7} \cmidrule(lr){8-10}
\textbf{Dataset} & Raw & Anat & Contrast & Raw & Anat & Contrast & Raw & Anat & Contrast \\\\
\midrule
OASIS & 0.92 & 0.93 & 0.71 & 0.86 & 0.81 & 0.71 & 5.15 & 5.45 & 6.28 \\
ADNI & 0.94 & 0.93 & 0.74 & 0.97 & 0.82 & 0.50 & 4.46 & 4.43 & 5.61 \\
HCP & 0.95 & 0.96 & 0.81 & 0.96 & 0.90 & 0.70 & 2.67 & 2.99 & 2.77 \\
ALL & 0.96 & 0.96 & 0.78 & 0.91 & 0.93 & 0.68 & 3.91 & 4.35 & 5.21 \\

\bottomrule
\end{tabular}}
\end{table*}

To determine whether demographic signals reflect biological variation or site-related acquisition-specific effects, we evaluate cross-dataset generalization by training on one dataset and testing on another. This introduces shifts in acquisition protocols, scanner vendors, and population characteristics, allowing us to assess whether demographic signals persist under distribution shift (Table~\ref{tab:cross_dataset_generalization}). Across all tasks, performance drops markedly under dataset shift, particularly for age prediction, where MAE increases dramatically (e.g., HCP$\rightarrow$ADNI: Raw 42.39 years). However, age-transfer results involving HCP should be interpreted in the context of the markedly different cohort age ranges. HCP participants span 23.5–36 years, whereas OASIS and ADNI participants span 42–97 and 55–95 years, respectively. These experiments therefore capture both acquisition-domain shift and a substantial shift in the age distribution. For classification, a consistent pattern emerges: anatomy-focused representations ($z_{\text{anat}}$) tend to generalize more robustly than raw images. For example, ADNI$\rightarrow$OASIS sex prediction improves from BalAcc 0.76 (Raw) to 0.84 (Anat). Whereas, contrast embeddings ($z_{\text{contrast}}$) frequently collapse toward chance (0.50), indicating poor transferability of acquisition-dependent signal. These findings suggest that while contrast embeddings encode measurable demographic information within a dataset, much of this signal is dataset-specific and does not generalize across sites. Structural representations, in contrast, retain more stable demographic information under distribution shift. Together with the within-dataset results, this indicates that demographic predictability in brain MRI arises mainly from anatomical sources, and acquisition-driven correlations are less robust and more site-dependent.

\begin{table*}[t!]
\centering
\caption{Cross-dataset generalization. \textbf{Sex} and \textbf{Race} are reported as balanced accuracy (BalAcc, $\uparrow$; equal class weighting), and \textbf{Age} is reported as mean absolute error (MAE, $\downarrow$).}
\label{tab:cross_dataset_generalization}
\setlength{\tabcolsep}{7pt}
\renewcommand{\arraystretch}{1.15}
\resizebox{\textwidth}{!}{
\begin{tabular}{llccccccccc}
\toprule
\textbf{Train} & \textbf{Test}
& \multicolumn{3}{c}{\textbf{Sex (BalAcc $\uparrow$)}}
& \multicolumn{3}{c}{\textbf{Race (BalAcc $\uparrow$)}}
& \multicolumn{3}{c}{\textbf{Age (MAE $\downarrow$)}} \\
\cmidrule(lr){3-5}\cmidrule(lr){6-8}\cmidrule(lr){9-11}
& & Raw & Anat & Contrast & Raw & Anat & Contrast & Raw & Anat & Contrast \\
\midrule

\multirow{2}{*}{\textbf{OASIS}} 
& HCP 
& 0.65 & 0.82 & 0.50 
& 0.85 & 0.87 & 0.50 
& 26.37 & 28.03 & 29.89 \\
& ADNI 
& 0.88 & 0.89 & 0.59 
& 0.93 & 0.99 & 0.53 
& 5.15 & 5.50 & 8.71 \\

\midrule

\multirow{2}{*}{\textbf{HCP}} 
& OASIS 
& 0.72 & 0.71 & 0.53 
& 0.85 & 0.82 & 0.70 
& 39.55 & 42.42 & 46.91 \\
& ADNI 
& 0.60 & 0.61 & 0.50 
& 0.88 & 0.84 & 0.60 
& 42.39 & 44.24 & 50.61 \\

\midrule

\multirow{2}{*}{\textbf{ADNI}} 
& OASIS 
& 0.76 & 0.84 & 0.70 
& 0.81 & 0.84 & 0.50 
& 5.17 & 5.49 & 6.47 \\
& HCP 
& 0.62 & 0.81 & 0.51 
& 0.86 & 0.90 & 0.50 
& 29.58 & 28.11 & 30.77 \\

\bottomrule
\end{tabular}}
\end{table*}

\begin{table*}[b!]
\centering
\caption{OASIS multi-sequence analysis. Performance across sequences for models trained on raw images ($f_{\text{full}}$), anatomy-focused representations ($f_{\text{anat}}$), and contrast-only embeddings ($f_{\text{contrast}}$). 
\textbf{Sex} and \textbf{Race} are reported as balanced accuracy (BalAcc, $\uparrow$; equal class weighting). \textbf{Age} is reported as mean absolute error (MAE, $\downarrow$).}
\label{tab:oasis_sequences}
\setlength{\tabcolsep}{7pt}
\renewcommand{\arraystretch}{1.15}
\resizebox{\textwidth}{!}{
\begin{tabular}{l ccc ccc ccc}
\toprule
& \multicolumn{3}{c}{\textbf{Sex (BalAcc $\uparrow$)}} 
& \multicolumn{3}{c}{\textbf{Race (BalAcc $\uparrow$)}} 
& \multicolumn{3}{c}{\textbf{Age (MAE $\downarrow$)}} \\
\cmidrule(lr){2-4} \cmidrule(lr){5-7} \cmidrule(lr){8-10}
\textbf{Sequence}
& Raw & Anat & Contrast 
& Raw & Anat & Contrast 
& Raw & Anat & Contrast \\
\midrule

T1w  
& 0.92 & 0.93 & 0.71 
& 0.86 & 0.81 & 0.71 
& 5.15 & 5.45 & 6.28 \\

T2w  
& 0.91 & 0.90 & 0.72 
& 0.74 & 0.88 & 0.51 
& 5.44 & 5.91 & 6.60 \\

FLAIR  
& 0.71 & 0.83 & 0.75 
& 0.65 & 0.85 & 0.50 
& 4.54 & 4.80 & 6.67 \\

\midrule

All Sequences  
& 0.91 & 0.93 & 0.76
& 0.84 & 0.81 & 0.70 
& 5.37 & 5.16 & 5.90 \\

\bottomrule
\end{tabular}}
\end{table*}

We further investigate the demographic signal across imaging sequences within OASIS (Table~\ref{tab:oasis_sequences}) to assess whether the relative roles of anatomy and contrast vary with sequence type. Across T2w and FLAIR images, anatomy representations remain close to raw performance (e.g., sex BalAcc 0.91 vs. 0.90 for T2w), indicating that structural variation continues to account for a substantial portion of the demographic signal beyond T1-weighted imaging. Contrast embeddings retain measurable predictive power across sequences. For example, sex prediction from $z_{\text{contrast}}$ reaches BalAcc 0.72 in T2w and 0.75 in FLAIR whereas race prediction remains near chance. Joint training on all sequences maintains strong sex prediction performance, suggesting that combining contrasts does not remove demographic signal. Anatomy representations continue to perform strongly (sex BalAcc 0.93, age MAE 5.16), while contrast embeddings remain weaker yet informative. Overall, these results show that demographic predictability extends beyond T1-weighted imaging where anatomy-focused representations largely contain the demographic signal across sequences and contrast embeddings contain a weaker, more sequence-dependent signal.

\section{Discussion}
In this study, we go beyond showing that demographic attributes can be predicted from brain MRI to address a more fundamental question: where does this signal originate? By decomposing MRI into anatomy-focused and contrast-dependent representations, our framework examines the relative contributions of anatomical structure and acquisition-dependent contrast to demographic predictability.

Our results consistently show that the majority of the demographic signal is preserved in anatomy-focused representations, establishing structural variation as the dominant contributor. This aligns with biological expectations for age and is plausible for sex, given known morphological differences \cite{agesexbrain,sexhandednesscerebral,sexdifferencesmeta}, and there is evidence that relative brain volume can differ across racial and ethnic groups~\cite{racebrain,corticalracebrain}. At the same time, contrast embeddings retain stable, above-chance predictive performance across datasets and sequences which might be due to population-level acquisition correlations or minor disentanglement leakage.

These findings carry important implications. Because a substantial portion of the demographic signal resides in anatomical structure, approaches that focus exclusively on acquisition harmonization or intensity normalization are unlikely to fully mitigate demographic predictability in downstream applications such as Alzheimer's disease detection. Conversely, strategies that indiscriminately remove anatomical variation risk suppressing clinically meaningful information. Effective bias mitigation should therefore focus on identifying and controlling the specific anatomical features that contribute to demographic signal while preserving those relevant to the clinical task. Acquisition-related variations should be addressed when they introduce additional demographic information, particularly in site-specific settings.

Identifying demographic signal does not in itself imply harmful bias in clinical predictions. However, understanding where this signal resides is a prerequisite for evaluating its downstream impact. Our framework provides a basis for investigating how these components propagate into clinical tasks and under what conditions they may translate into measurable bias, supporting the development of causally informed and fair neuroimaging models.

\subsection{Limitations}
\label{sec:limitations}

The image-based predictors are based on 3D ResNet-50, whereas the contrast embeddings are evaluated using a smaller MLP because the inputs differ in dimensionality and structure. Consequently, performance differences may partly reflect predictor capacity and should not be interpreted as a pure measure of the information contained in each representation. Analysis on the extent to which this affects the interpretations is left for future work.

Another limitation of this study is the imbalanced racial composition of the included datasets, particularly ADNI, where only 84 participants (5.2\%) are Black. Although balanced accuracy mitigates the effect of class imbalance on the reported metrics, the small minority-class size limits the statistical reliability of race prediction results. Future work should evaluate these effects in more demographically representative cohorts.

    

\begin{credits}

\subsubsection{\ackname}
This research was supported by the UK Engineering and Physical Sciences Research Council (EPSRC) [grant reference number EP/Y035216/1] Centre for Doctoral Training in Data-Driven Health (DRIVE-Health) at King's College London, with additional support from deepc GmbH. Further support was provided by the Scientific and Technological Research Council of T\"{u}rkiye (T\"{U}B\.{I}TAK) under the 2213-A Overseas Graduate Scholarship. Data used in preparation of this article were obtained from the Alzheimer's Disease Neuroimaging Initiative (ADNI) database (\url{https://adni.loni.usc.edu}). As such, the investigators within ADNI contributed to the design and implementation of ADNI and/or provided data but did not participate in the analysis or writing of this report. A complete listing of ADNI investigators can be found at \url{https://adni.loni.usc.edu/wp-content/uploads/how_to_apply/ADNI_Acknowledgement_List.pdf}. Data were also provided in part by OASIS and the Human Connectome Project, WU-Minn Consortium (Principal Investigators: David Van Essen and Kamil Ugurbil; 1U54MH091657), funded by the 16 NIH Institutes and Centers that support the NIH Blueprint for Neuroscience Research, and by the McDonnell Center for Systems Neuroscience at Washington University.

\subsubsection{\discintname}
The authors have no competing interests to declare that are relevant to the content of this article.
\end{credits}


%
%
%

\bibliographystyle{splncs04}
\bibliography{contrast}

@incollection{avci2026mrclip,
  author       = {Avci, M. Yigit and Borges, Pedro and Wright, Paul and Yigitsoy, Mert and Ourselin, Sebastien and Cardoso, Jorge},
  title        = {{MR-CLIP}: Efficient Metadata-Guided Learning of {MRI} Contrast Representations},
  booktitle    = {Machine Learning in Medical Imaging. MLMI 2025},
  editor       = {Cui, Zhiyong and Rekik, Ammar and Suk, Heung-Il and Ouyang, Xiaoxiao and Sun, Ke and Wang, Shuo},
  series       = {Lecture Notes in Computer Science},
  volume       = {16241},
  publisher    = {Springer, Cham},
  year         = {2026},

}

@misc{avci2025metadataaligned3dmrirepresentations,
      title={Metadata-Aligned {3D MRI} Representations for Contrast Understanding and Quality Control}, 
      author={Mehmet Yigit Avci and Pedro Borges and Virginia Fernandez and Paul Wright and Mehmet Yigitsoy and Sebastien Ourselin and Jorge Cardoso},
      year={2025},
      eprint={2511.00681},
      archivePrefix={arXiv},
      primaryClass={cs.CV},
     howpublished={\href{http://arxiv.org/abs/2511.00681}{arXiv:2511.00681}},
}

@article{synthstrip,
    title = {{SynthStrip}: skull-stripping for any brain image},
    journal = {{NeuroImage}},
    volume = {260},
    pages = {119474},
    year = {2022},
    author = {Andrew Hoopes and Jocelyn S. Mora and Adrian V. Dalca and Bruce Fischl and Malte Hoffmann},
}

@article{oasis,
	author = {LaMontagne, Pamela J. and Benzinger, Tammie LS. and Morris, John C. and Keefe, Sarah and Hornbeck, Russ and Xiong, Chengjie and Grant, Elizabeth and Hassenstab, Jason and Moulder, Krista and Vlassenko, Andrei G. and Raichle, Marcus E. and Cruchaga, Carlos and Marcus, Daniel},
	title = {{OASIS-3}: Longitudinal Neuroimaging, Clinical, and Cognitive Dataset for Normal Aging and {Alzheimer Disease}},
	elocation-id = {2019.12.13.19014902},
	year = {2019},
	publisher = {Cold Spring Harbor Laboratory Press},
	journal = {medRxiv}
}

@article{Lee_2025,
   title={Understanding-informed Bias Mitigation for Fair {CMR} Segmentation},
   volume={3},
   ISSN={2766-905X},
   journal={Machine Learning for Biomedical Imaging},
   publisher={Machine Learning for Biomedical Imaging},
   author={Lee, Tiarna and Puyol-Anton, Esther and Ruijsink, Bram and Masci, Pier-Giorgio and Keehn, Louise and Chowienczyk, Phil and Haseler, Emily and Shi, Miaojing and King, Andrew},
   year={2025},
 pages={808–824} }

@article{gichoya2022ai,
  title        = {{AI} recognition of patient race in medical imaging: a modelling study},
  author       = {Gichoya, Judy Wawira and Banerjee, Imon and Bhimireddy, Ananth Reddy and Burns, John L and Celi, Leo Anthony and Chen, Li-Ching and Correa, Ramon and Dullerud, Natalie and Ghassemi, Marzyeh and Huang, Shih-Cheng and Kuo, Po-Chih and Lungren, Matthew P and Palmer, Lyle J and Price, Brandon J and Purkayastha, Saptarshi and Pyrros, Ayis T and Oakden-Rayner, Lauren and Okechukwu, Chima and Seyyed-Kalantari, Laleh and Trivedi, Hari and Wang, Ryan and Zaiman, Zachary and Zhang, Haoran},
  journal      = {The Lancet Digital Health},
  volume       = {4},
  number       = {6},
  pages        = {e406--e414},
  year         = {2022},
  publisher    = {Elsevier},
}

@article{lotter2024acquisition,
  author    = {Lotter, William},
  title     = {Acquisition parameters influence {AI} recognition of race in chest {X}-rays and mitigating these factors reduces underdiagnosis bias},
  journal   = {Nature Communications},
  volume    = {15},
  pages     = {7465},
  year      = {2024},
}

@article{Underdiagnosis_2021, title={Underdiagnosis bias of artificial intelligence algorithms applied to chest radiographs in under-served patient populations}, author={Seyyed-Kalantari, Laleh and Zhang, Haoran and McDermott, Matthew and Chen, Irene and Marzyeh, Ghassemi}, journal={Nature Medicine}, volume={27}, pages={2176–2182}, year={2021} }

@article{
larrazabal2020,
author = {Agostina J. Larrazabal  and Nicolás Nieto  and Victoria Peterson  and Diego H. Milone  and Enzo Ferrante },
title = {Gender imbalance in medical imaging datasets produces biased classifiers for computer-aided diagnosis},
journal = {Proceedings of the NAS},
volume = {117},
number = {23},
pages = {12592-12594},
year = {2020},

}

@misc{schrouff2023diagnosingfailuresfairnesstransfer,
      title={Diagnosing failures of fairness transfer across distribution shift in real-world medical settings}, 
      author={Jessica Schrouff and Natalie Harris and Oluwasanmi Koyejo and Ibrahim Alabdulmohsin and Eva Schnider and Krista Opsahl-Ong and Alex Brown and Subhrajit Roy and Diana Mincu and Christina Chen and Awa Dieng and Yuan Liu and Vivek Natarajan and Alan Karthikesalingam and Katherine Heller and Silvia Chiappa and Alexander D'Amour},
      year={2023},
      eprint={2202.01034},
      archivePrefix={arXiv},
      primaryClass={cs.LG},
      url={https://arxiv.org/abs/2202.01034}, 
}

@misc{avci2025distcliparbitrarymetadataimage,
  title={{DIST-CLIP}: Arbitrary Metadata and Image Guided {MRI} Harmonization via Disentangled Anatomy-Contrast Representations}, 
      author={Mehmet Yigit Avci and Pedro Borges and Virginia Fernandez and Paul Wright and Mehmet Yigitsoy and Sebastien Ourselin and Jorge Cardoso},
      year={2025},
      eprint={2512.07674},
      archivePrefix={arXiv},
      primaryClass={cs.CV},
      url={https://arxiv.org/abs/2512.07674}, 
}

@misc{achara2025invisibleattributesvisiblebiases,
      title={Invisible Attributes, Visible Biases: Exploring Demographic Shortcuts in MRI-based Alzheimer's Disease Classification}, 
      author={Akshit Achara and Esther Puyol Anton and Alexander Hammers and Andrew P. King},
      year={2025},
      eprint={2509.09558},
      archivePrefix={arXiv},
      primaryClass={cs.CV},
      url={https://arxiv.org/abs/2509.09558}, 
}

@article{lamontagne2019oasis3,
  title={OASIS-3: Longitudinal Neuroimaging, Clinical, and Cognitive Dataset for Normal Aging and Alzheimer Disease},
  author={LaMontagne, Pamela J. and Benzinger, Tammie L.S. and Morris, John C. and Keefe, Sarah and Hornbeck, Russ and Xiong, Chengjie and Grant, Elizabeth and Hassenstab, Jason and Moulder, Krista and Vlassenko, Andrei and Raichle, Marcus E. and Cruchaga, Carlos and Marcus, Daniel},
  journal={medRxiv},
  year={2019},
}

@article{hcp,
title = {{The WU-Minn Human Connectome Project: An overview}},
journal = {NeuroImage},
volume = {80},
pages = {62-79},
year = {2013},
issn = {1053-8119},
author = {David C. {Van Essen} and Stephen M. Smith and Deanna M. Barch and Timothy E.J. Behrens and Essa Yacoub and Kamil Ugurbil},
}

@article{jack2008adni_mri,
  title={The {A}lzheimer's {D}isease {N}euroimaging {I}nitiative {(ADNI)}: {MRI} methods},
  author={Jack, Clifford R. Jr. and Bernstein, Matthew A. and Fox, Nick C. and Thompson, Paul and Alexander, Gene and Harvey, Danielle and Borowski, Barbara and Britson, Paul J. and Whitwell, Jennifer L. and Ward, Chad and Dale, Anders M. and Felmlee, Joel P. and Gunter, Jeffrey L. and Hill, Derek L.G. and Killiany, Ronald and Schuff, Norbert and Fox-Bosetti, Susan and Lin, Ching-Po and Studholme, Colin and DeCarli, Charles S. and Krueger, Gunnar and Ward, Heidi A. and Metzger, Gregory J. and Scott, Kathleen T. and Mallozzi, Ronald and Blezek, Daniel and Levy, Jonathan and Debbins, John P. and Fleisher, Adam S. and Albert, Marilyn and Green, Robert and Bartzokis, George and Glover, Gary and Mugler, John and Weiner, Michael W.},
  journal={Journal of Magnetic Resonance Imaging},
  volume={27},
  number={4},
  pages={685--691},
  year={2008},
  pmid={18302232},
  pmcid={PMC2544629}
}

@article{biasinrad,
author = {Waite, Stephen and Scott, Jinel and Colombo, Daria},
title = {Narrowing the Gap: Imaging Disparities in Radiology},
journal = {Radiology},
volume = {299},
number = {1},
pages = {27-35},
year = {2021},
note ={PMID: 33560191},
}

@InProceedings{biasmri2022,
author={Petersen, Eike
and Feragen, Aasa
and da Costa Zemsch, Maria Luise
and Henriksen, Anders
and Wiese Christensen, Oskar Eiler
and Ganz, Melanie},
editor={Wang, Linwei
and Dou, Qi
and Fletcher, P. Thomas
and Speidel, Stefanie
and Li, Shuo},
title={Feature Robustness and Sex Differences in Medical Imaging: A Case Study in {MRI}-Based {Alzheimer's Disease} Detection},
booktitle={Medical Image Computing and Computer Assisted Intervention -- MICCAI 2022},
year={2022},
publisher={Springer Nature Switzerland},
address={Cham},
pages={88--98},
isbn={978-3-031-16431-6},
}

@misc{he2015deepresiduallearningimage,
      title={Deep Residual Learning for Image Recognition}, 
      author={Kaiming He and Xiangyu Zhang and Shaoqing Ren and Jian Sun},
      year={2015},
      eprint={1512.03385},
      archivePrefix={arXiv},
      primaryClass={cs.CV},
      url={https://arxiv.org/abs/1512.03385}, 
}

@InProceedings{esthercmrfairness,
author="Puyol-Ant{\'o}n, Esther
and Ruijsink, Bram
and Piechnik, Stefan K.
and Neubauer, Stefan
and Petersen, Steffen E.
and Razavi, Reza
and King, Andrew P.",
editor="de Bruijne, Marleen
and Cattin, Philippe C.
and Cotin, St{\'e}phane
and Padoy, Nicolas
and Speidel, Stefanie
and Zheng, Yefeng
and Essert, Caroline",
title="Fairness in Cardiac MR Image Analysis: An Investigation of Bias Due to Data Imbalance in Deep Learning Based Segmentation",
booktitle="Medical Image Computing and Computer Assisted Intervention -- MICCAI 2021",
year="2021",
publisher="Springer International Publishing",
address="Cham",
pages="413--423",
isbn="978-3-030-87199-4"
}

@article{melba:2025:035:danaee,
    title = "Investigating Demographic Bias in Brain MRI Segmentation: A Comparative Study of Deep-Learning and Non-Deep-Learning Methods ",
    author = "Danaee, Ghazal and Niethammer, Marc and Rushmore, Jarrett and Bouix, Sylvain",
    journal = "Machine Learning for Biomedical Imaging",
    volume = "3",
    issue = "Special issue on FAIMI",
    year = "2025",
    pages = "792--808",
    issn = "2766-905X",
}

@article{agesexbrain,
title = {Age and sex effects on brain morphology},
journal = {Progress in Neuro-Psychopharmacology and Biological Psychiatry},
volume = {21},
number = {8},
pages = {1231-1237},
year = {1997},
issn = {0278-5846},
author = {Theodore J. Passe and Pradeep Rajagopalan and Larry A. Tupler and Christopher E. Byrum and James R. Macfall and K.Ranga Rama Krishnan}
}

@article{racebrain,
    author = {Brickman, Adam M. and Schupf, Nicole and Manly, Jennifer J. and Luchsinger, José A. and Andrews, Howard and Tang, Ming X. and Reitz, Christiane and Small, Scott A. and Mayeux, Richard and DeCarli, Charles and Brown, Truman R.},
    title = {Brain Morphology in Older African Americans, Caribbean Hispanics, and Whites From Northern Manhattan},
    journal = {Archives of Neurology},
    volume = {65},
    number = {8},
    pages = {1053-1061},
    year = {2008},
    month = {08},

    issn = {0003-9942},
    
}

@article{sexhandednesscerebral,
  title={Cerebral asymmetry and the effects of sex and handedness on brain structure: a voxel-based morphometric analysis of 465 normal adult human brains},
  author={Good, Catriona D and Johnsrude, Ingrid and Ashburner, John and Henson, Richard NA and Friston, Karl J and Frackowiak, Richard SJ},
  journal={Neuroimage},
  volume={14},
  number={3},
  pages={685--700},
  year={2001},
  publisher={Elsevier}
}

@article{sexdifferencesmeta,
  title={A meta-analysis of sex differences in human brain structure},
  author={Ruigrok, Amber NV and Salimi-Khorshidi, Gholamreza and Lai, Meng-Chuan and Baron-Cohen, Simon and Lombardo, Michael V and Tait, Roger J and Suckling, John},
  journal={Neuroscience \& Biobehavioral Reviews},
  volume={39},
  pages={34--50},
  year={2014},
  publisher={Elsevier}
}

@article{corticalracebrain,
  title={Differences in cortical structure between cognitively normal East Asian and Caucasian older adults: a surface-based morphometry study},
  author={Kang, Dong Woo and Wang, Sheng-Min and Na, Hae-Ran and Park, Sonya Youngju and Kim, Nak Young and Lee, Chang Uk and Kim, Donghyeon and Son, Seong-Jin and Lim, Hyun Kook},
  journal={Scientific reports},
  volume={10},
  number={1},
  pages={20905},
  year={2020},
  publisher={Nature Publishing Group UK London}
}

@article{solovyev20223d,
  title={3D convolutional neural networks for stalled brain capillary detection},
  author={Solovyev, Roman and Kalinin, Alexandr A and Gabruseva, Tatiana},
  journal={Computers in Biology and Medicine},
  volume={141},
  pages={105089},
  year={2022},
  publisher={Elsevier},
}

@article{synthseg,
    title = {{SynthSeg}: Segmentation of brain {MRI} scans of any contrast and resolution without retraining},
    journal = {Medical Image Analysis},
    volume = {86},
    pages = {102789},
    year = {2023},
    author = {Benjamin Billot and Douglas N. Greve and Oula Puonti and Axel Thielscher and Koen Van Leemput and Bruce Fischl and Adrian V. Dalca and Juan Eugenio Iglesias},
}

\end{document}